\documentclass[letterpaper, 10 pt, conference]{ieeeconf}  

\IEEEoverridecommandlockouts                              
\usepackage[utf8]{inputenc}
\usepackage[T1]{fontenc}
\usepackage[acronym]{glossaries}
\usepackage{paralist}
\usepackage{graphicx}
\usepackage{array,multirow}
\usepackage[T1]{fontenc}

\title{\LARGE \bf
Generative \acrshort{ai} and Teachers - For Us or Against Us? A Case Study
}

\author{Jenny Pettersson, Elias Hult, Tim Eriksson
 and
Tosin Adewumi$^{1}$
\thanks{We thank Björn Backe for his support for this work. This work was partially supported by the Wallenberg AI, Autonomous Systems and Software Program (WASP), funded by the Knut and Alice Wallenberg Foundation and counterpart funding from Luleå University of Technology (LTU).}
\thanks{$^{1}$Corresponding author
        {\tt\small firstname.lastname@ltu.se}}%
\\
Machine Learning Group, EISLAB, Luleå University of Technology, Sweden}

\makeglossaries

\newacronym{ml}{ML}{Machine Learning}
\newacronym{llm}{LLM}{Large Language Model}
\newacronym{sota}{SotA}{state-of-the-art}
\newacronym{iaa}{IAA}{inter-annotator agreement}
\newacronym{nlp}{NLP}{natural language processing}
\newacronym{ai}{AI}{artificial intelligence}
\newacronym{llama}{LLaMA}{Large Language Model Meta AI}
\newacronym{bert}{BERT}{Bidirectional Encoder Representations from Transformers}
\newacronym{svm}{SVM}{support vector machine}
\newacronym{gru}{GRU}{gated recurrent unit}
\newacronym{cnn}{CNN}{convolutional neural network}
\newacronym{t5}{T5}{Text-to-Text Transfer Transformer}
\newacronym{gai}{GenAI}{generative artificial intelligence}
\newacronym{ltu}{LTU}{Luleå University of Technology}
\newacronym{unesco}{UNESCO}{United Nations Educational, Scientific and Cultural Organization}

\begin{document}

\maketitle
\thispagestyle{empty}
\pagestyle{empty}

\begin{abstract}

We present insightful results of a survey on the adoption of \acrfull{gai} by university teachers in their teaching activities.
The transformation of education by \acrshort{gai}, particularly large language models (\acrshort{llm}s), has been presenting both opportunities and challenges, including cheating by students.
We prepared the online survey according to best practices and the questions were created by the authors, who have pedagogy experience.
The survey contained 12 questions and a pilot study was first conducted.
The survey was then sent to all teachers in multiple departments across different campuses of the university of interest in Sweden: Luleå University of Technology.
The survey was available in both Swedish and English.
The results show that 35 teachers (more than half) use \acrshort{gai} out of 67 respondents.
Preparation is the teaching activity with the most frequency that \acrshort{gai} is used for and
ChatGPT is the most commonly used \acrshort{gai}.
59\% say it has impacted their teaching, however, 55\% say there should be legislation around the use of \acrshort{gai}, especially as inaccuracies and cheating are the biggest concerns.

\end{abstract}

\section{INTRODUCTION}
\label{intro}

Recent advances in \acrfull{ai}, especially \acrfull{gai}, have caused a stir in the Education sector around the world \cite{adewumi2023procot,alasadi2023generative}.
ChatGPT\footnote{chat.openai.com/}, the leading \acrfull{llm} by OpenAI, has both been beneficial and controversial.
Some of the concerns with \acrshort{gai} are generating deepfakes and the nature of these "unexplainable" models \cite{holmes2023guidance}.
In spite of these concerns, many recognize the benefits inherent in these technologies \cite{chan2023ai,holmes2023guidance}.
In this case study, we seek to understand how university teachers perceive \acrshort{gai} and investigate the following research questions.
\begin{enumerate}
    \item To what extent are university teachers open to adopting generative AI in their teaching and classrooms?
    \item What is the correlation between the impact of \acrshort{gai} on teachers' teaching activities and their encouragement of their students to use it?
\end{enumerate}

There's increasing study of the impact of \acrshort{gai} on students \cite{chan2023ai,holmes2023guidance}.
It is equally important to study the impact on other stakeholders or teachers' teaching activities in Education \cite{dai2023reconceptualizing}.
\acrfull{unesco} hopes \acrshort{gai} will be a tool that benefits teachers, students, and researchers.
Our main \textbf{contributions} include the following: (1) We show through data that university teachers in this case study are open to adopting \acrshort{gai}; (2) we demonstrate the correlation between the positive impact of \acrshort{gai} on teachers to their willingness to encourage their students to adopt it; (3) we provide many qualitative examples of comments of teachers on the impact of \acrshort{gai}, ways they encourage students, and their concerns.

The rest of this paper is organised as follows.
The literature review is discussed in Section \ref{litrev}.
The method employed in this work is described in detail in Section \ref{method}.
The findings are discussed in Section \ref{results_sec}.
We conclude with closing remarks in Section \ref{conclude}.

\section{Literature Review}
\label{litrev}

The subject of \acrshort{gai} in teaching is gaining increasing attention.
It's impact on pedagogy cannot be ignored.
Recent \acrshort{llm}s, such as ChatGPT, Aurora-M \cite{Nakamura2024AuroraMTF}, \acrfull{llama}-2 \cite{touvron2023llama} and a host of others, have compelling abilities to generate human-like content, based on their training with big data \cite{holmes2023guidance}.
This has prompted \acrfull{unesco} to publish the guidance for \acrshort{gai} in education and research \cite{holmes2023guidance}, which builds on their recommendation on the ethics of \acrshort{ai}.
To gauge the awareness of teachers or educators and their adoption of \acrshort{gai}, it is useful to conduct a survey, similarly to that done with students \cite{chan2023students,smolansky2023educator}.

\cite{krosnick2018questionnaire} identified some of the best practices for designing the questionnaire for a survey.
As simple as it may sound, the ordering of questions is an important consideration \cite{jensen1999s,kreuter2011effects}.
These best practices are essential to have quality data from the survey.
\cite{krosnick2018questionnaire} also show that it is important to design questions to avoid \textit{acquiescence} bias, which is the endorsement of a statement, regardless of the content.
Testing a survey in a pilot study usually improves the quality of the full survey, as emphasized by \cite{oksenberg1991new,presser2004methods,baburajan2020open,bowden2002methods}.
Using either closed or open questions have their own benefits, as demonstrated by \cite{schuman1987problems,reja2003open,connor2019comparing}.
In our work, we combined both types of questions to get the best our of the survey.

\section{Method}
\label{method}

Our chosen method for evaluating the use of \acrshort{gai} by teachers at \acrshort{ltu} is through a survey.
It was designed to gather information about their habitual use of \acrshort{gai} in teaching.
In creating the survey, a few factors were taken into consideration to keep the inquiry objective and to avoid phrasings that could potentially skew the answers of the participants, based on best practices \cite{krosnick2018questionnaire} .
The factors are given below but are not limited to them.
\begin{itemize}
    \item Keep the questions simple and concise, so as not to produce off-topic answers \cite{gehlbach2018survey,artino2012last}.
    
    \item Include a broad, exhaustive list of viewpoints \cite{krosnick2018questionnaire} 

    \item Avoid questions where people tend to agree or disagree with statements regardless of their actual feelings or beliefs \cite{krosnick2018questionnaire}

    \item Avoid single or double negation questions \cite{krosnick2018questionnaire}.
\end{itemize}

To reach a larger target audience, the survey questionnaire was created in both Swedish and English using an online tool.\footnote{Miscrosoft Forms}
The Swedish translation was carried out by some of the authors of this work, who are native speakers.
The translations in the Results and Appendix sections were machine-translated and vetted by the native speakers.
The survey consists of 12 questions to cover a relatively broad range of concepts that are necessary to answer the research questions.
For the purpose of the survey, a simplified definition was provided for \acrshort{gai}: \textit{Generative AI, such as ChatGPT or DALL-E, is a tool that can answer questions and create images and other media based on prompts from the user} \cite{holmes2023guidance}.
A copy of the questionnaire is available online.\footnote{forms.office.com/e/jvrmPPaJJh}

The survey was anonymous and a pilot study involving 8 teachers was initially carried out to ascertain if the forms and questions needed any adjustments.
This full study involved sending the online survey via a link to all the teachers in multiple departments in different geographical campuses of \acrshort{ltu}.
The following are the 12 questions in the survey and their answer options, as created by the authors based on their pedagogy experience and the factors mentioned earlier.
\textit{Q1} and \textit{Q8} were multiple choice questions.
For teachers who answer ``\textit{None}'' to the first question ("\textit{filter question}" \cite{krosnick2018questionnaire}), they were directed to continue to question 6 onwards.

\begin{enumerate}
    \item \textit{Have you used any generative AI in any of your teaching activities (e.g. preparation, teaching, assessment, or none)?}\\
  \begin{inparaenum}[i)]
    \item Preparation
    \item Teaching
    \item Assessment
    \item Research\footnote{Undersökning in Swedish, though initially translated as Undervisning.}
    \item Administration
    \item None
    \item Other
  \end{inparaenum}

    \item \textit{ Which ones? e.g GenAI, Ex. ChatGPT, DALL-E, Bing AI, Google Bard etc. Others }
    \item \textit{How often do you use the one you use most?}
  \begin{inparaenum}[i)]
    \item Once a month
    \item Once a week
    \item Twice or more a week
    \item Less than once a month
  \end{inparaenum}
    
    \item \textit{Do you think the use has impacted your teaching?}
  \begin{inparaenum}[i)]
    \item Yes
    \item No
    \item Not sure
  \end{inparaenum}
    
    \item \textit{Briefly describe the impact on your teaching.}
    
    \item \textit{Do you think AI will replace teachers in your subject if the trend of AI development continues?}
  \begin{inparaenum}[i)]
    \item Yes
    \item No
    \item Not sure
  \end{inparaenum}
    
    \item \textit{Do you think there should be legislation around the use of generative AI?}
  \begin{inparaenum}[i)]
    \item Yes
    \item No
    \item Ambivalent
  \end{inparaenum}
    
    \item \textit{What are some of your ethical concerns about generative AI?}
  \begin{inparaenum}[i)]
    \item Gender bias
    \item Racial bias
    \item Inaccuracies
    \item Cheating
    \item None
    \item Other concerns
  \end{inparaenum}
    
    \item \textit{Will you encourage any of your students to use generative AI (in an ethical manner)?}
  \begin{inparaenum}[i)]
    \item Yes
    \item No
  \end{inparaenum}
    
    \item \textit{If you answered “Yes” in the previous question, In what way? And if "No", please say why.}
    
    \item \textit{Your gender}
  \begin{inparaenum}[i)]
    \item Woman
    \item Man
    \item Non-binary
    \item Prefer not to say
  \end{inparaenum}
    
    \item \textit{Your Division and Department}
\end{enumerate}

The following \acrshort{ltu} departments were involved in filling the survey.

\begin{enumerate}
    \item SRT: The Department of Computer Science, Electrical and Space Engineering, among other subjects, contains Pervasive and Mobile Computing, Digital Services and Systems, Computer Science, Signals and Systems, Robotics and AI, Space Technology, Cyber-Physical Systems and Machine Learning.
    
    \item SBN: The Civil, Environmental and Natural Resources Engineering department, among other subjects, contains Urban Water Engineering, Architecture, Structural Engineering, Building Materials, Engineering Acoustics, Soil Mechanics, Ore Geology, Applied Geophysics, Applied Geochemistry, Chemical Technology, and Process Metallurgy
    
    \item ETKS: The Department of Social Sciences, Technology and Arts, among other subjects, contains Industrial Marketing, Political Science, Human Work Sciences, Performing Arts, Musical Performance, Economics, and Design.
    
    \item HLT: The Department of Health, Education and Technology, among other subjects, contains Occupational Therapy, English and Education, Physiotherapy, Biomedical Engineering, Medical Science, Nursing, and Psychology.
    
    \item TVM: The Department of Engineering Sciences and Mathematics, among other education, contains Mechanical Engineering, Automotive Engineering, Sustainable Energy, Electrical Power, Engineering Physics, and Mechanical Engineering.

\end{enumerate}

\section{Results and Discussion}
\label{results_sec}

\begin{figure}[h!]
\centering
\includegraphics[width=0.5\textwidth]{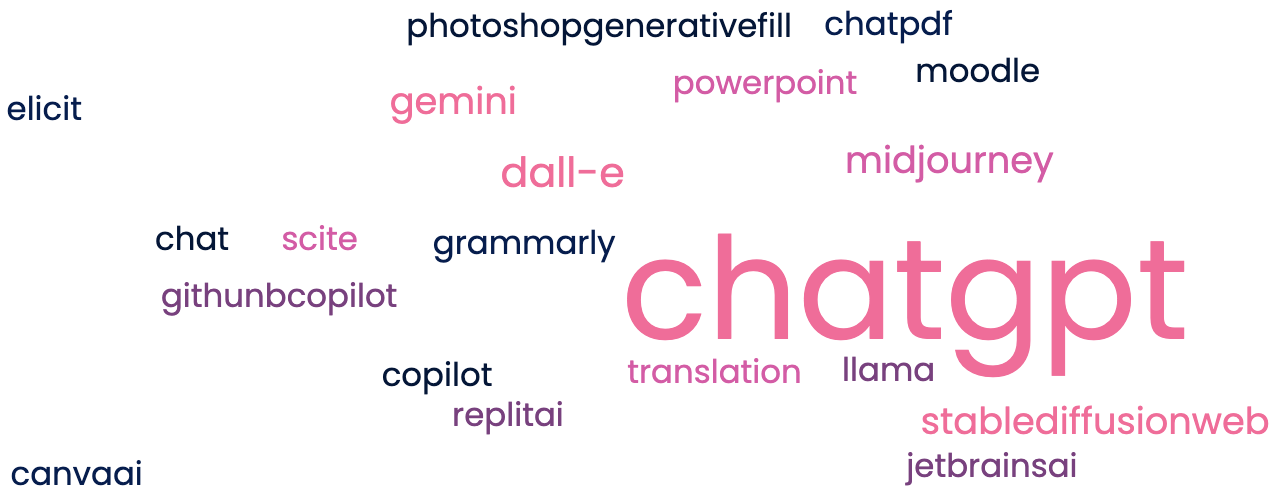}
\caption{WordCloud of \acrshort{gai}} \label{figwc}
\end{figure}

\begin{figure*}[h]
\centering
\includegraphics[width=0.8\textwidth]{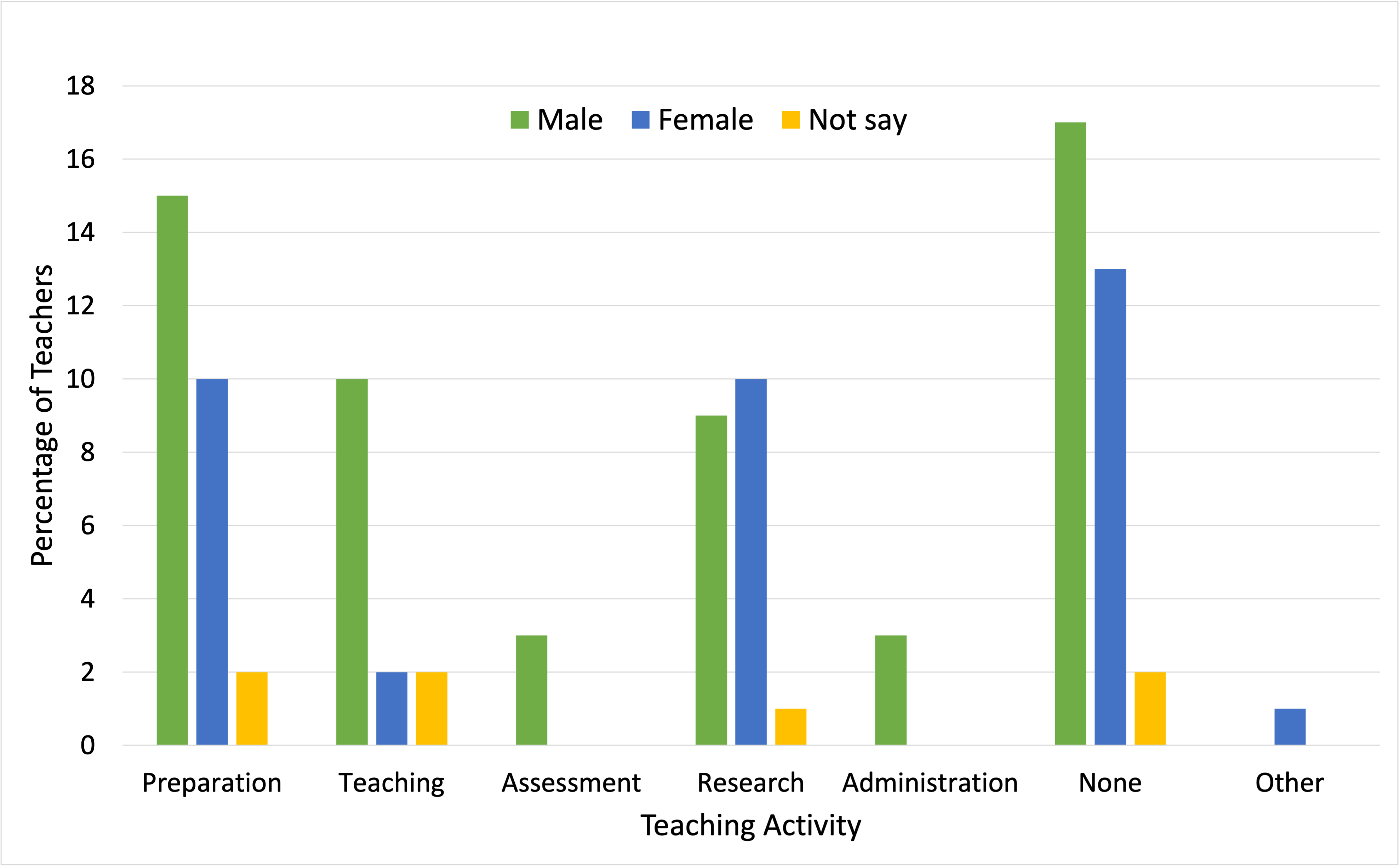}
\caption{\acrshort{gai} usage activities across gender out of a total percentage of 100\%.} \label{figa}
\end{figure*}

The survey took 3:53 minutes for each teacher to complete on average.
From Table \ref{tresults}, 32 (48\%) of the teachers do not use \acrshort{gai} in any of their teaching activities.
The remaining 35 (52\%) use \acrshort{gai} for one or more teaching activities, where \textit{Preparation} is the most frequent activity, being 27\%.
Besides the identified teaching activities, one teacher explained that she uses \acrshort{gai} "\textit{To see if student work is AI generated}".
The wordcloud of Figure \ref{figwc} shows that ChatGPT has the lion share of usage with 52\% of frequency of mentions (26 out of 50), in answer to \textit{Q2}.
The nearest is DALL-E, with 6\%.
Gemini, Stable Diffusion Web, and Midjourney come in next at 4\% while all the other \acrshort{gai}s have 2\% (only 1 mention).
Figure \ref{figa} shows the distribution of activities across gender.

\begin{table*}[h!]
\centering
\caption{Results in percentage (\%). The gender and department sections each add up to the total.}\label{tresults}
\begin{tabular}{| p{0.08\linewidth} | p{0.1\linewidth} | p{0.04\linewidth} || p{0.03\linewidth} | p{0.03\linewidth} | p{0.03\linewidth} || p{0.03\linewidth} | p{0.03\linewidth} | p{0.04\linewidth} | p{0.03\linewidth} | p{0.03\linewidth} |}
\hline
\textbf{Question} & \textbf{Option} & \textbf{Total}  & \multicolumn{2}{c}{\textbf{Gender}} & & \multicolumn{4}{c}{\textbf{Department}} &\\\cline{4-11} 
 &  & & \textbf{M} & \textbf{W}  & \textbf{Not say}  & \textbf{SRT} & \textbf{SBN} & \textbf{ETKS} & \textbf{HLT} & \textbf{TVM} \\
\hline
\multirow{3}{*}{Q1} & Preparation & 27 & 15 & 10 & 2 & 3.6 & 3.6 & 14.4 & 3.6 & 1.8 \\
 & Teaching & 14 & 10 & 2 & 2 & 6 & 2 & 6 & 0 & 0 \\
 & Assessment & 3 & 3 & 0 & 0 & 0 & 0& 0& 0& 3\\
& Research& 20 & 9 & 10 & 1 & 7.5 & 5 & 7.5 & 0 & 0 \\
& Administration & 3 & 3 & 0 & 0 & 3 & 0 & 0& 0& 0\\
& None & 32 & 17 & 13 & 2 & 9.41 & 7.53 & 9.41 & 5.65 & 0\\
& Other & 1 & 0 & 1 & 0 & 0& 1 & 0& 0& 0\\
  & &  & & &  & & & & &  \\
\hline
\multirow{3}{*}{Q3} &  Once a month & 18.42  & 13.16 & 5.26 & 0 & 3.07 & 6.14 & 6.14 & 3.07 & 0 \\
 & Once a week & 18.42 & 10.53 & 7.89 & 0 & 9.21 & 0 & 9.21 &0  &0 \\
& > once a week & 31.58 & 15.79 & 7.89 & 7.89 & 21.05 & 0 & 5.26 & 0 & 5.26\\
& < once a month & 31.58 & 15.79 & 15.79 & 0 & 3.95 & 11.84 & 11.84 & 3.95 & 0\\
  & &  & & & &  & & & & \\
\hline
\multirow{3}{*}{Q4} &  Yes & 58.97 & 41.03 & 12.82 & 5.13 & 19.66 & 14.74 & 19.66 & 0 & 4.91\\
 & No & 17.95 & 10.26 & 7.69 & 0 & 0 & 8.97 & 4.49  & 4.49 & 0 \\
  & Not sure & 23.08 & 5.13 & 15.38 & 2.56 & 11.54 & 0 & 7.69 & 3.85 & 0\\
  & &  & & &  & & & & & \\
\hline
\multirow{3}{*}{Q6} &  Yes & 4.48 & 4.48 & 0 & 0 & 4.48 & 0& 0& 0& 0 \\
 & No & 85.07 & 41.79 & 35.82 & 7.46 & 29.10 & 15.67 & 29.10 & 11.19 & 0\\
  & Not sure & 10.45 & 8.96 & 1.49 & 0 & 3.48 & 3.48 & 1.74 & 0 & 1.74\\
  & &  & & & & & & & &  \\
\hline
\multirow{3}{*}{Q7} &  Yes & 55.22 & 29.85 & 19.40 & 5.97 &  23.36 & 12.74 & 14.87 & 2.12 & 2.12\\
 & No & 17.91 & 11.94 & 4.48 & 1.49 & 5.12 & 5.12 & 7.67 & 0 & 0 \\
& Ambivalent & 26.87 & 13.43 & 13.43 & 0 & 6.72 & 2.24 & 6.72 & 11.19 & 0 \\
  & &  & & &  & & & & & \\
\hline
\multirow{3}{*}{Q8} & Gender bias & 15.34 & 7.98 & 5.52 & 1.84 & 5.41 & 1.80 & 6.32 & 0.90 & 0.90\\
 & Racial bias & 15.34 & 7.36 & 6.13 & 1.84 & 5.96 & 1.70 & 5.96 & 0.85 & 0.85 \\
 & Inaccuracies & 34.36 & 18.40 & 13.50 & 2.45 & 11.75 & 7.23 & 10.85 & 3.62 & 0.90 \\
& Cheating & 26.99 & 15.34 & 10.43 & 1.23 & 8.70 & 4.35 & 7.84 & 5.22 & 0.87 \\
& None & 1.23 & 1.23 & 0 & 0 & 1.23 & 0 & 0 & 0 & 0 \\
& Other & 6.75 & 2.45 & 3.68 & 0.61 & 1.12 & 2.25 & 3.37 & 0 & 0\\
  & &  & & &  & & & & & \\
\hline
\multirow{3}{*}{Q9} & Yes & 76.12 & 44.78 & 26.87 & 4.48 & 30.45 & 17.40 & 23.92 & 2.17 & 2.17\\
 & No & 23.88 & 10.45 & 10.45 & 2.99 &  4.34 & 2.17 & 6.51 & 10.85 & 0\\
  & &  & & & &  & & & & \\
\hline
\multirow{3}{*}{Q11} & Woman& 37.31  \\
 & Man & 55.22 \\
  & Non-binary & 0 \\
  & Prefer not say & 7.46 \\\cline{0-2}
\end{tabular}
\end{table*}

Of the 52\% who use these tools, there are as many teachers who use them twice or more in a week (i.e. more than once) as there are those who use them less than once a month (31.58\%).
About 59\% say \acrshort{gai} has impacted their teaching.
Indeed, we see a strong positive correlation between the \textit{Yes} choices of the departments to \textit{Q4} (\textit{Do you think the use has impacted your teaching?}) and \textit{Q9} (\textit{Will you encourage any of your students to use generative AI}?), based on the Spearman's correlation coefficient (\(\rho\)) +0.9474,\footnote{r = 1 and -1 are perfect positive and negative correlations, respectively} for \textit{p} = 0.01438 (2-tailed).
This implies the association between the two variables can be considered statistically significant.
Several examples of the ways teachers feel \acrshort{gai} has impacted their teaching are given in the appendix.

\begin{figure*}[h]
\centering
\includegraphics[width=0.8\textwidth]{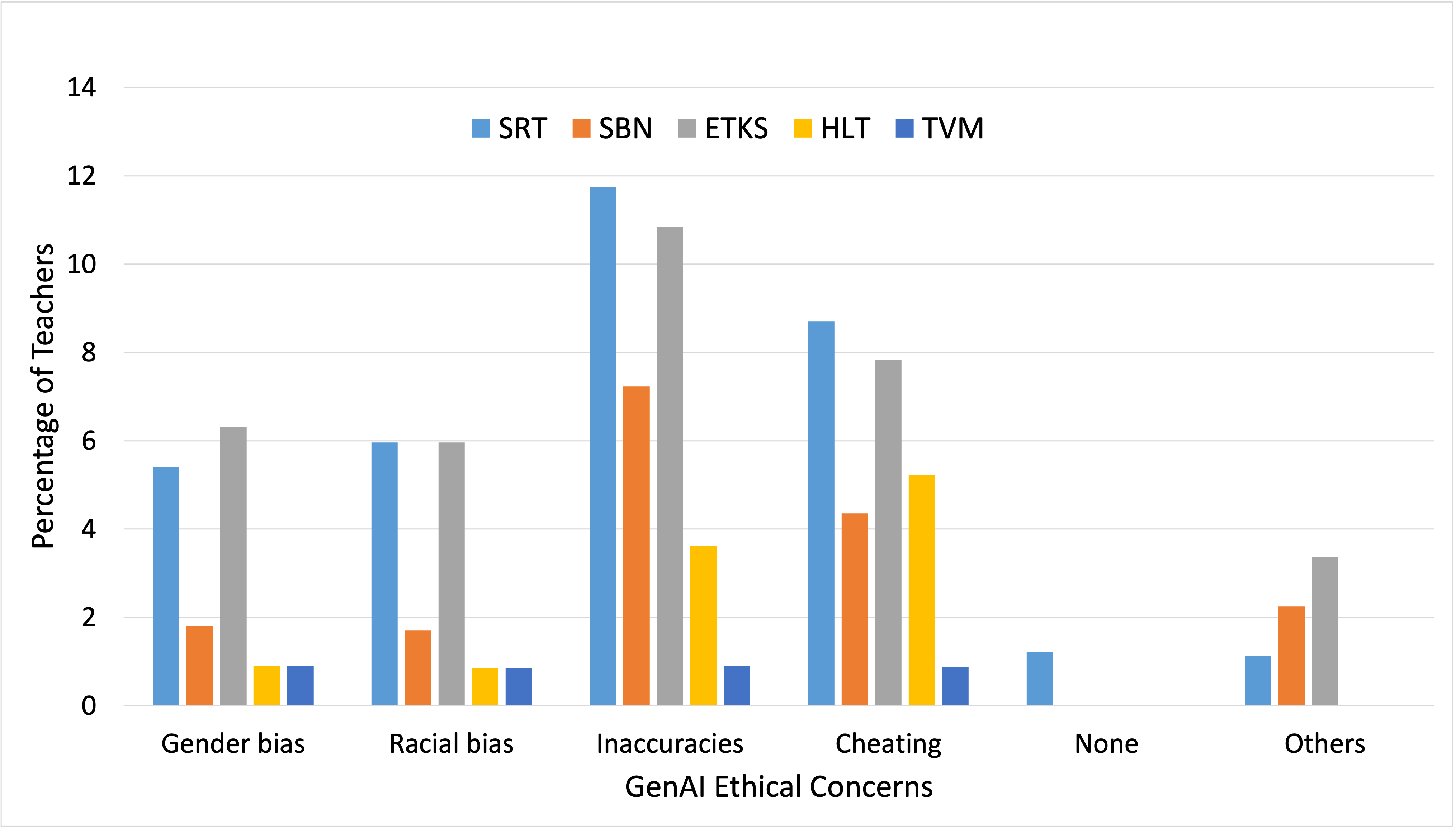}
\caption{\acrshort{gai} concerns across departments out of a total percentage of 100\%.} \label{figc}
\end{figure*}

Most (85\%) of the 67 teachers do not think \acrshort{ai} will replace teachers.
Many (55\%) say there should be legislation around the use of \acrshort{gai}, possibly because of the ethical concerns observed.
Inaccuracies (34.36\%) and cheating (26.99\%) are the two most common concerns teachers have.
Figure \ref{figc} shows the concerns across the departments.
Given that the majority of teachers encourage their students to use \acrshort{gai}, some of the ways they go about it are listed in the appendix, including these:

\begin{itemize}
    \item "\textit{...I do 1 or 2 sessions on how and why they should use it. I also show them when it can give wrong results and how to fact check it...}"
    \item "\textit{seed texts, help in checking texts and results; use AI as study buddy}"
    \item "\textit{Det är ett mycket effektivt hjälpmedel och bör uppmuntras. Är man orolig för fusk så examinerar man studenterna på fel sätt. (It is a very effective aid and should be encouraged. If you are worried about cheating, you are examining the students in the wrong way.)}"
    
\end{itemize}
Overall, 37.31\%, 55.22\%, and 7.46\% of the teachers who completed the survey were women, men and those who preferred not to say.

\section{Conclusion}
\label{conclude}

We have shown in this case study that teachers are open to adopting \acrshort{gai}, as over 50\% currently use it.
We also observe a strong positive correlation between the positive impact of \acrshort{gai} on their teaching activities and their willingness to encourage their students to adopt \acrshort{gai}.
We agree with the comments of some of the teachers that students "\textit{are guaranteed to use it already}", therefore we believe teachers should be knowledgeable about these tools in order to provide the appropriate guidance for students.
Future work can investigate some of the ways of addressing the concerns of teachers expressed in this study.

\section*{ACKNOWLEDGMENT}

We deeply appreciate all the teachers who completed the survey.
Their valuable response made this work possible.
We thank Björn Backe for his support towards this work.
This work was partially supported by the Wallenberg AI, Autonomous Systems and Software Program (WASP), funded by the Knut and Alice Wallenberg Foundation and counterpart funding from Luleå University of Technology (LTU).

\section*{APPENDIX}
\label{appendix}

\subsection*{Examples by teachers of impact on teaching}
\begin{enumerate}
    \item I am able to provide more bang for the bucks and student feedback shows the impact.
    \item I made my subject knowledge deeper.
    \item ChatGPT hjälper mig att sammanfatta innehållet i t ex en workshop eller en föreläsning (\textit{ChatGPT helps me summarize the content of, for example, a workshop or a lecture})
    \item t.ex. bättre bilder till mitt undervisningsmaterial, hjälp med bra översättning till engelska etc.	(\textit{for example better pictures for my teaching material, help with a good translation into English, etc.})
    \item preparation of slides and text material is more efficient and result more impactful.
    \item Bättre språk, fler exempel (\textit{Better language, more examples})
    \item I show its use for students to use properly as a tool, as well as when or how not to use it.
    \item improved clarity
    \\
    \item Jag får ett bollplank som kan hjälpa mig som lärare. (\textit{I get a sounding board that can help me as a teacher}.)
    \item Jag har använt Canva:s generativ AI för att snabbt ta fram illustrationer till mina powerpoints. Jag tror att det kan ha en viss positiv påverkan för inlärning att få "bildstöd" till anteckningarna. (\textit{I have used Canva's generative AI to quickly produce illustrations for my powerpoints. I think having "visual support" for the notes can have some positive impact on learning}.)
    \item Tidsbesparande (\textit{Timesaving})
    \item Ibland dyker det upp aspekter som jag tidigare inte tänkt på, men som är relevanta. (\textit{Sometimes aspects appear that I previously did not think about, but which are relevant.})
    \item Är ett fantastiskt verktyg att skapa bilder istället för att leta clip-art. Att få hjälp att förklara saker samt som kreativt verktyg i idegenereringsprocessen	(\textit{Is a great tool to create images instead of looking for clip-art. To get help explaining things and as a creative tool in the idea generation process})
    \item Jag har fått en bättre förståelse för hur studenter kan använda det som stöd samt jag har lärt mig att känna igen resultaten i studenters arbete (\textit{I have gained a better understanding of how students can use it as support and I have learned to recognize the results of students' work})
    
\end{enumerate}

\subsection*{Examples of other concerns}
\begin{enumerate}
    \item I have had trouble with students copying AI generated information. I am afraid they are not using it as a learning tool, but rather to avoid learning.
    \item ...students' usage affects the type of tasks I can give them and how I test their knowledge. That's mainly why I use it myself.
    \item Nya former för examination kräva (\textit{New forms of examination require})
    \item Efter att tidigare ha använt hemtentor i delar av kursen har jag gått över till salskrivningar. (\textit{Having previously used take-home exams in parts of the course, I have switched to classroom writing.})
    \item Jag har tydligt kunnat visa för mina studenter varför det är viktigt att kunna ha grundläggande kunskap inom ett område, för det AI säger behöver inte nödvändigtvis vara korrekt, vilket de fick erfara i en kurs. (I have been able to clearly show my students why it is important to be able to have basic knowledge in a field, because what the AI says does not necessarily have to be correct, as they experienced in a course.)
    \item I sin nuvarande form är generativ AI bra på att generera text som ser rimlig ut men mycket väl kan vara ful av felaktigheter. Jag ser inte detta som särskilt anändbart för mina studenter. (\textit{In its current form, generative AI is good at generating text that looks reasonable but may well be ugly with inaccuracies. I don't see this as particularly relevant to my students.})
\end{enumerate}

\subsection*{Examples of ways teachers encourage their students to use \acrshort{gai}}

\begin{enumerate}
    \item "Use it to learn, not to cheat". Använd för att förbättra eget material, inte för att generera från grunden. Viktigt att man inte presenterar andras material som sitt eget. Däremot är det liten skillnad att få en språkfgransking av en människa eller från AI när man väl skrivit texten. Viktigt att kunna materialet så att man kan faktagranska AI-lösningarna. (\textit{"Use it to learn, not to cheat". Use to improve your own material, not to generate from scratch. It is important not to present other people's material as your own. However, there is little difference in getting a language check by a human or from AI once you have written the text. It is important to know the material so that you can fact-check the AI solutions.})

    \item Jag uppmuntrar dem att använda AI så mycket som möjligt om det hjälper deras lärande. (\textit{I encourage them to use AI as much as possible if it helps their learning.})
    \item T.ex. för att bolla idéer, få hjälp med struktur i en text, hitta och sortera källor. Jag tycker det är ett väldig kraftfylld verktyg men precis som vilken verktyg som helst kan den vara farlig om den används av människor utan rätt kunskap. Så att skaffa sig just den kunskapen för att kunna använda generativ AI på ett säkert sätt bor ingår undervisningen. (\textit{For example. to brainstorm ideas, get help with structure in a text, find and sort sources. I think it's a very powerful tool but like any tool it can be dangerous if used by people without the right knowledge. So acquiring that particular knowledge to be able to use generative AI in a safe way is part of the teaching.})
    \item De använder det garanterat redan, så bättre att ha riktlinjer kring hur användandet bör ske. (\textit{They are guaranteed to use it already, so better to have guidelines about how the use should take place.})
    \item Inte uppmuntra, men inte heller hindra (\textit{Not encouraging, but not hindering either})
        \\
    \item Framförallt för att lära sig skriva vetenskaplig text på engelska och för att diskutera kursinnehåll, mjukvarukunskaper (t.ex. Excel, Matlab, python etc.) (\textit{Mainly to learn how to write scientific text in English and to discuss course content, software skills (e.g. Excel, Matlab, python etc.)})
    \item Studenterna (Och industrin) använder redan generativ AI i långt högre utsträckning än vad vi lärare gör. Bättre att lära från dem och uppmana dem till att använda systemen på ett klokt sätt som uppmuntrar deras lärande. (\textit{Students (And industry) already use generative AI to a far greater extent than we teachers do. Better to learn from them and encourage them to use the systems wisely which encourages their learning.})
    \item Precis som jag skrev innan så har jag uppmuntrat studenterna att använda ChatGPT för att få förståelse/fördjupning av vissa ämnen (\textit{Just as I wrote before, I have encouraged the students to use ChatGPT to gain understanding/deepening of certain topics})
    \item Leta material, sortera i material (\textit{Find materials, sort in materials})
    \item ja, att använda det i den kreativa processen för att utforska en mängd idéer, för att förbättra engelskan i texter etc. (\textit{yes, to use it in the creative process to explore a variety of ideas, to improve English in texts, etc.})
    \\
    \item Som ett skrivstöd	(\textit{As a writing aid})
    
    \item För språkgranskning och programmering (\textit{For language review and programming})
    \item AI kan vara mycket hjälpfullt om man har ett koncept/begrepp som man inte förstår men vill ha förklarat så att man sen, självständigt, kan använda konceptet/begreppet. (\textit{AI can be very helpful if you have a concept that you don't understand but want explained so that you can then, independently, use the concept.})
    \item Utmärkt för sammanfattningar och scanning av stora litteraturmängder (\textit{Excellent for summaries and scanning large volumes of literature})
    \item utkast till texter, utkast till musik, tex. (\textit{drafts of texts, drafts of music, e.g.})
    \\
    \item Det är ett ypperligt bollplank, framförallt då man inte har någon fysisk person att diskutera med, men det kan även ge förstärkning om man använder den då man är studerar i grupp. (\textit{It is an excellent sounding board, especially when you have no physical person to discuss with, but it can also provide reinforcement if you use it when you are studying in a group.})
    \item I produktutveckling, som ett stöd och verktyg. (\textit{In product development, as a support and tool.})
    \item skriv hjälp, hjälp att komma igång med ett arbete, skapa bilder för presentationer, hjälp mot skrivkramp (\textit{write help, help to get started with a work, create images for presentations, help against writing cramp})
    \item 1. Använda det för att stava rätt i inlämningar. 2. För att ha någon att bolla idéer kring studentarbete. 3. Fråga om grundläggande koncept som AI kan behärska och förklara. (\textit{1. Use it to spell correctly in submissions. 2. To have someone to bounce ideas off of student work. 3. Ask about basic concepts that AI can master and explain.})
    \item Vi lärare behöver skapa och formulera premisser som är rimliga för detta, nu när man inte kan backa bandet med AI. Hur - vet jag inte än. (\textit{We teachers need to create and formulate premises that are reasonable for this, now that you cannot reverse the trend with AI. How - I don't know yet.})

\end{enumerate}

\bibliographystyle{acm}
\bibliography{ref}

\end{document}